\documentclass[12pt,a4paper]{article}
\usepackage{multirow}
\usepackage{amsmath}
\usepackage{graphicx}
\usepackage{tabularx}
\usepackage{booktabs}
\usepackage{array}
\usepackage[colorlinks=true,
    linkcolor=black,
    citecolor=black,
    urlcolor=black]{hyperref}
\usepackage{cite}
\usepackage{geometry}
\usepackage{tikz}
\usetikzlibrary{calc}
\usepackage{chngcntr}
\usepackage{float}  
\counterwithin{figure}{section}
\usepackage{tikz}
\usepackage{booktabs}
\usetikzlibrary{shapes.geometric, arrows.meta, positioning, calc, shadows, fit}

\title{\textbf{A-MBER: Affective Memory Benchmark for Emotion Recognition}}
 
\begin{document}
\begin{titlepage}
\thispagestyle{empty} 
\begin{tikzpicture}[remember picture,overlay]
    \node at (current page.center) {
        \includegraphics[width=\paperwidth,height=\paperheight]{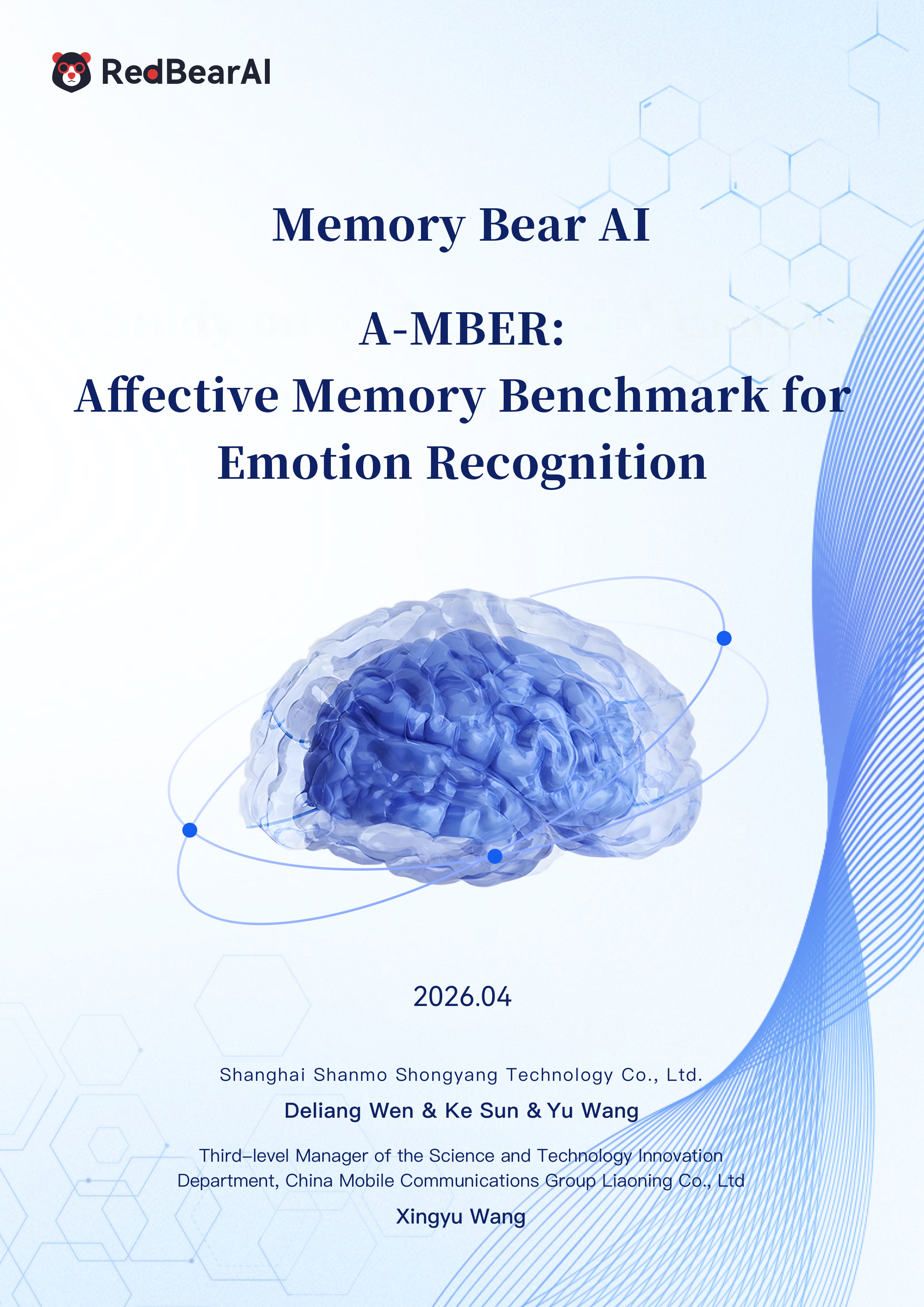}
    };
\end{tikzpicture}
\end{titlepage}


\begin{abstract}
AI assistants that interact with users over time need to interpret the user's current emotional state in order to provide appropriate and personalized responses. A user's present emotional state may depend on earlier events, recurring triggers, prior support attempts, and longer emotional trajectories that unfold across sessions. Yet existing resources still separate this problem into two disconnected evaluation settings: emotion datasets mainly assess local or instantaneous affect, while long-term memory benchmarks focus largely on factual recall, temporal consistency, or knowledge updating. This leaves limited support for evaluating whether a model can use remembered interaction history to interpret a user's present affective state.

We introduce \textbf{A-MBER}, an \textbf{Affective Memory Benchmark for Emotion Recognition}, to evaluate this missing capability. A-MBER targets \emph{present affective interpretation grounded in remembered multi-session interaction history}. Given an interaction trajectory and a designated anchor turn, a model must infer the user's current affective state, identify the historically relevant evidence, and justify its interpretation in a grounded way. To support this goal, we construct the benchmark through a staged pipeline with explicit intermediate representations, including long-horizon planning, conversation generation, annotation, question construction, and final benchmark packaging. The resulting benchmark supports \emph{judgment}, \emph{retrieval}, and \emph{explanation} tasks, together with controlled robustness settings such as modality degradation and insufficient-evidence conditions.

Experiments compare local-context, long-context, retrieved-memory, structured-memory, and gold-evidence conditions within a unified evaluation framework. Results show that A-MBER is most discriminative on the subsets it is designed to stress, including long-range implicit affect, high-dependency memory levels, trajectory-based reasoning, and adversarial conditions. These findings suggest that memory contributes to affective interpretation not merely by exposing more historical content, but by enabling more selective, grounded, and context-sensitive use of interaction history.
\end{abstract}

\newpage
\tableofcontents

\newpage
\section{Introduction}

As AI systems move toward longer-term interaction, they increasingly need to infer a user's current affective state in order to respond appropriately, provide support, and sustain personalized behavior over time. This need arises in a wide range of settings, including tutoring, companionship, counseling, customer support, and other forms of repeated interaction. In such scenarios, the practical value of memory lies not only in storing more user history, but in using that history to support better understanding of the user in the present moment. This perspective is especially relevant for memory-centered systems such as Memory Bear, which frame long-term interaction as a memory-dependent process and emphasize adaptive memory maintenance, affect-sensitive interpretation, and more continuous user understanding across sessions \cite{wen2025memorybear,wen2026memoryscience}.

In many realistic cases, however, a user's present affective state cannot be inferred from the current turn alone. A brief, polite, or seemingly neutral response may in fact reflect disappointment, guardedness, suppression, or relational distance when interpreted against earlier events, recurring triggers, prior support attempts, or a longer emotional trajectory. The capability required in such cases is therefore neither generic long-term factual memory nor conventional local emotion recognition. Rather, it is the ability to use remembered multi-session interaction history to support \emph{present affective interpretation}: not merely assigning a local emotion label to the current turn, but inferring how the user's present state should be understood in light of prior interaction and grounding that interpretation in historically relevant evidence.

Existing evaluation resources do not capture this problem well. On one side, long-term conversational memory benchmarks have substantially advanced the evaluation of factual recall, temporal reasoning, knowledge updates, and related memory operations over extended interaction history \cite{maharana-etal-2024-evaluating,wu2024longmemeval}. On the other side, emotion and multimodal conversation datasets have advanced local emotion recognition, sentiment analysis, empathetic response, and emotion-cause grounding. What remains insufficiently studied is whether a model can use remembered multi-session interaction history to interpret a user's present affective state. In other words, the practical need for this capability is clear, but benchmark support for measuring it remains limited.

\begin{figure}[t]
\centering
\includegraphics[width=0.96\linewidth]{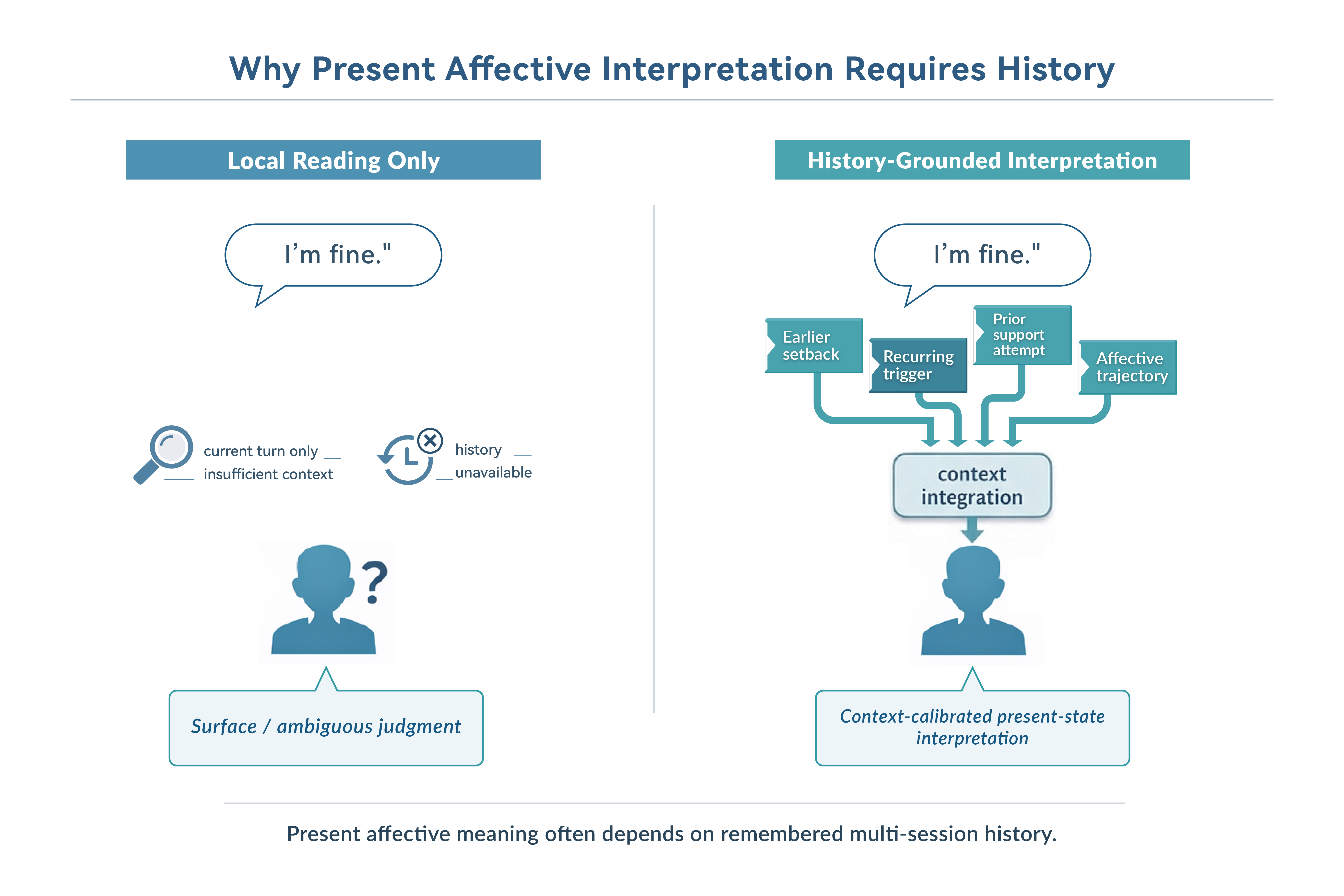}
\caption{Motivating contrast between local reading and history-grounded present affect interpretation. The figure illustrates why the same anchor turn may appear neutral or weakly emotional in isolation, yet become interpretable once earlier events, recurring triggers, and interaction trajectory are taken into account.}
\label{fig:motivation_contrast}
\end{figure}

To address this problem, and as Figure~\ref{fig:motivation_contrast} makes concrete, we introduce \textbf{A-MBER}, an \textbf{Affective Memory Benchmark for Emotion Recognition}. The name is intended to foreground the benchmark's focus on affective memory while remaining close in spirit to the broader Memory Bear line of memory-centered research. A-MBER is designed to evaluate whether a model can use multi-session interaction history to interpret the user's present affective state at a designated anchor turn, rather than relying on the local turn alone. Given an interaction trajectory, an anchor turn, and a task condition, the model must infer the user's current affective state, identify the historically relevant evidence, and explain why that interpretation is warranted. The goal is therefore not only to test whether a model reaches a plausible affective judgment, but whether it reaches that judgment for the right historical reasons.

To support this goal, we construct A-MBER through a staged benchmark pipeline with explicit intermediate representations. In the present study, the primary construction route is a single-agent pipeline, chosen for controllability, schema stability, and benchmark production at scale. At the same time, the broader framework also accommodates agent-to-agent interaction as an auxiliary construction regime and leaves room for more deployment-oriented interaction settings in future extensions. What remains stable across these construction routes is the benchmark definition itself: a shared interaction representation, anchor-turn interface, evidence-grounding constraint, and task structure.

The benchmark is built around multi-session conversational interaction in which affective interpretation may depend on both textual content and delivery-related cues. In the present version, this information is represented through dialogue text together with structured delivery descriptions that preserve affect-relevant vocal properties in a controlled benchmark format. This design keeps the benchmark usable across model classes while retaining information that would otherwise be lost in plain transcript-only evaluation. It also makes it possible to test whether historically grounded interpretation remains stable when part of the locally available signal becomes weaker, ambiguous, or unavailable. Richer multimodal extensions, including stronger audio-grounded settings and additional perceptual channels, remain natural directions for future versions.

A-MBER is organized around three primary task families: \emph{judgment}, \emph{retrieval}, and \emph{explanation}. Judgment tasks ask whether the model can infer the correct present affective interpretation at the anchor turn. Retrieval tasks ask whether it can identify the historically relevant supporting turns. Explanation tasks ask whether it can justify the interpretation in a historically grounded way. In addition, the benchmark includes control, robustness-oriented, and adversarial settings designed to separate genuine long-horizon affective understanding from superficial local success.

Our contributions are threefold. First, we formulate \emph{affective memory for emotion recognition} as a benchmark target distinct from both generic long-term factual memory and conventional local emotion recognition. Second, we introduce a structured benchmark construction framework for this target, centered on staged generation and grounded intermediate representations. Third, we provide a benchmark setting that supports controlled evaluation across task family, memory dependency, reasoning structure, and robustness condition, making A-MBER suitable for studying how memory supports affective interpretation over time.

\section{Related Work}

\subsection{Long-Term Conversational Memory Benchmarks}

Recent work has begun to evaluate language agents on long-term conversational memory rather than only on single-session context use. LoCoMo introduces a benchmark for very long-term conversational memory, using a structured pipeline to construct multi-session conversations and evaluate models through question answering, event summarization, and dialogue generation \cite{maharana-etal-2024-evaluating}. LongMemEval similarly focuses on long-term interactive memory for chat assistants, evaluating abilities such as information extraction, multi-session reasoning, temporal reasoning, knowledge updates, and abstention \cite{wu2024longmemeval}. More recently, LoCoMo-Plus argues that conversational memory evaluation should move beyond explicit factual recall toward cognitively meaningful and implicitly constrained memory use \cite{li2026locomoplus}.

These benchmarks establish long-horizon interaction history as an important evaluation target, but their main sources of difficulty remain factual retention, temporal consistency, knowledge updating, or implicit constraint satisfaction. They are therefore highly relevant to our work, yet they do not directly evaluate whether a model can use remembered interaction history to interpret a user's present affective state.

\subsection{Emotion and Multimodal Conversation Datasets}

A separate line of work has developed datasets for emotion recognition, sentiment analysis, empathy, and explanation in dialogue. EmotionLines provides an early conversation-level emotion corpus with utterance-level emotion labels in multi-party dialogues \cite{hsu-etal-2018-emotionlines}. MELD extends this direction to multimodal multi-party dialogue by incorporating textual, acoustic, and visual signals \cite{poria-etal-2019-meld}. EmpatheticDialogues shifts the focus from emotion labeling to empathetic response generation in emotionally grounded conversations \cite{rashkin-etal-2019-towards}. RECCON further emphasizes interpretability by introducing emotion-cause recognition in conversation \cite{poria2021reccon}. Related multimodal affect resources such as CMU-MOSEI, together with widely used speech emotion datasets such as IEMOCAP, are also central resources for multimodal or acoustic affect analysis \cite{zadeh2018cmu-mosei,busso2008iemocap}.

These datasets are important foundations for affective computing and conversational affect modeling, but they address a different problem from the one considered here. Most focus on local emotion recognition, empathetic response, sentiment prediction, or emotion-cause grounding at the utterance or short-context level. They do not directly test whether a model can rely on remembered multi-session interaction history to interpret the user's present state.

\subsection{Synthetic Benchmark Construction and Memory-Centered Systems}

Synthetic and semi-synthetic pipelines have become increasingly important for building large, structured evaluation resources. LoCoMo, for example, uses a machine--human pipeline to create long-horizon conversational data and derive benchmark tasks from them \cite{maharana-etal-2024-evaluating}. More broadly, staged generation has shown the practical value of explicit intermediate representations for improving controllability, coverage, and consistency in data construction \cite{wang2023selfinstruct}. Multi-agent generation and verification frameworks further suggest that distributing generation and checking roles across multiple agents may improve diversity or realism in some settings, although such approaches can also reduce production stability under strong structural constraints \cite{sengupta2024magv}.

This line of work is closely related to our construction strategy. A-MBER is also built through staged benchmark construction, with structured intermediate layers used to maintain alignment between long-horizon planning, observable interaction, evidence grounding, and final task packaging. At the same time, the broader motivation of our benchmark is closely connected to memory-centered systems such as Memory Bear, which motivate the need for evaluation resources that go beyond factual recall and make affect-sensitive use of remembered history measurable under controlled conditions \cite{wen2025memorybear,wen2026memoryscience}. In this sense, our work is aligned with memory-centered system design, but its contribution is at the benchmark level: the goal is to provide an evaluation resource for long-horizon affective memory rather than to propose a new memory architecture.

\subsection{Positioning Against Existing Long-Term Memory Benchmarks}

Table~\ref{tab:longterm_benchmark_comparison} situates A-MBER relative to representative long-term conversational memory benchmarks. The purpose of the comparison is not to rank benchmarks by scale alone, but to clarify the difference in evaluation target and design emphasis.

Compared with LoCoMo and LongMemEval, A-MBER is centered on present affective interpretation rather than factual recall, temporal reasoning, or knowledge updating as primary targets. Compared with LoCoMo-Plus, which pushes conversational memory evaluation toward beyond-factual cognitive constraints, A-MBER focuses more specifically on \emph{affective memory}: whether a model can use remembered multi-session interaction history to interpret the user's present state in a historically grounded way. In this setting, long-horizon difficulty is defined not only by raw turn count or text length, but also by the density of event dependencies, affect-relevant history, and cross-session interpretive burden. This makes it possible to evaluate emotionally and relationally meaningful long-range reasoning even when interaction length is not maximized in the same way as in some very-long-context benchmarks.

\begin{figure}[t]
\centering
\includegraphics[width=0.92\linewidth]{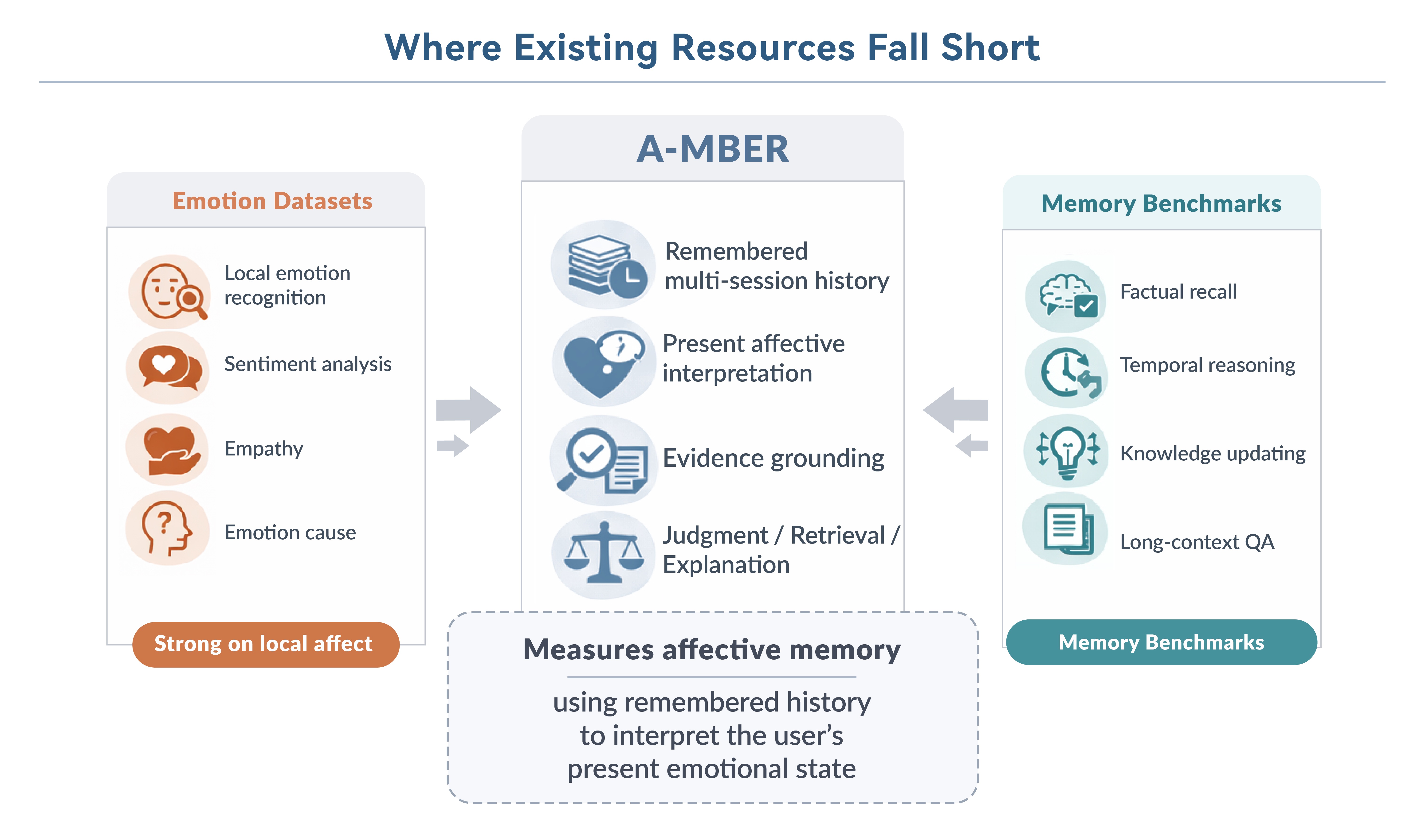}
\caption{Positioning of A-MBER relative to existing evaluation spaces. The figure visualizes the gap between resources that mainly test local or instantaneous emotion understanding and benchmarks that emphasize long-term memory for factual or temporal reasoning, locating A-MBER at their intersection around history-grounded present affect interpretation.}
\label{fig:gap_positioning}
\end{figure}

Table~\ref{tab:longterm_benchmark_comparison} provides a more detailed benchmark-level comparison, while Figure~\ref{fig:gap_positioning} offers a higher-level view of the same niche by placing A-MBER between local or instantaneous emotion evaluation and long-term memory benchmarks centered on factual or temporal reasoning.

\begin{table}[H]
\centering
\small
\setlength{\tabcolsep}{4pt}
\renewcommand{\arraystretch}{1.1}
\begin{tabularx}{\linewidth}{l>{\centering\arraybackslash}X>{\centering\arraybackslash}X>{\centering\arraybackslash}X>{\centering\arraybackslash}X}
\toprule
\textbf{Dimension} & \textbf{LongMemEval} & \textbf{LoCoMo} & \textbf{LoCoMo-Plus} & \textbf{A-MBER} \\
\midrule
Target & Interactive memory & Conversational memory & Cognitive memory & Affective memory \\
Present-state interpretation & Limited & Limited & Yes & Yes \\
History beyond local cues & Partial & Partial & Yes & Yes \\
Affective trajectory & No & No & Partial & Yes \\
Main difficulty & Memory operations & Long-span retention & Implicit constraints & Affective event logic \\
\bottomrule
\end{tabularx}
\caption{Comparison with representative long-term memory benchmarks.}
\label{tab:longterm_benchmark_comparison}
\end{table}

\section{Benchmark Construction Framework}

This section describes the construction framework of A-MBER. We first define the benchmark target and task interface, then describe the scenario and data representation used in the benchmark, followed by the primary construction pipeline adopted in this study. We finally discuss alternative interaction realizations, including agent-to-agent interaction as an auxiliary construction regime and hardware-mediated interaction as a possible extension path.

\subsection{Benchmark Target and Task Interface}

A-MBER targets \emph{affective memory} in long-horizon interaction: the ability to use remembered multi-session interaction history to support present affective interpretation. Relative to conventional long-term memory evaluation, the difference lies not only in what must be remembered, but in how remembered history must be used. Retrieval remains necessary, but it is not sufficient. The model must identify which prior events, interaction patterns, or relational cues are affectively relevant, connect them to the anchor turn, and justify the resulting interpretation through observable evidence. The main source of difficulty therefore lies less in raw history length alone than in implicitness, underdetermination, and dependence on the right historical context.

Operationally, each benchmark item is organized around a designated \emph{anchor turn}. Given a multi-session interaction history, an anchor turn, and a task condition, the model is asked to produce a task-specific output tied to that moment. This anchor-turn formulation keeps evaluation focused on present-state interpretation rather than global dialogue summarization and ensures that benchmark targets remain grounded in observable interaction rather than hidden planning variables.

\begin{figure}[t]
\centering
\includegraphics[width=0.88\linewidth]{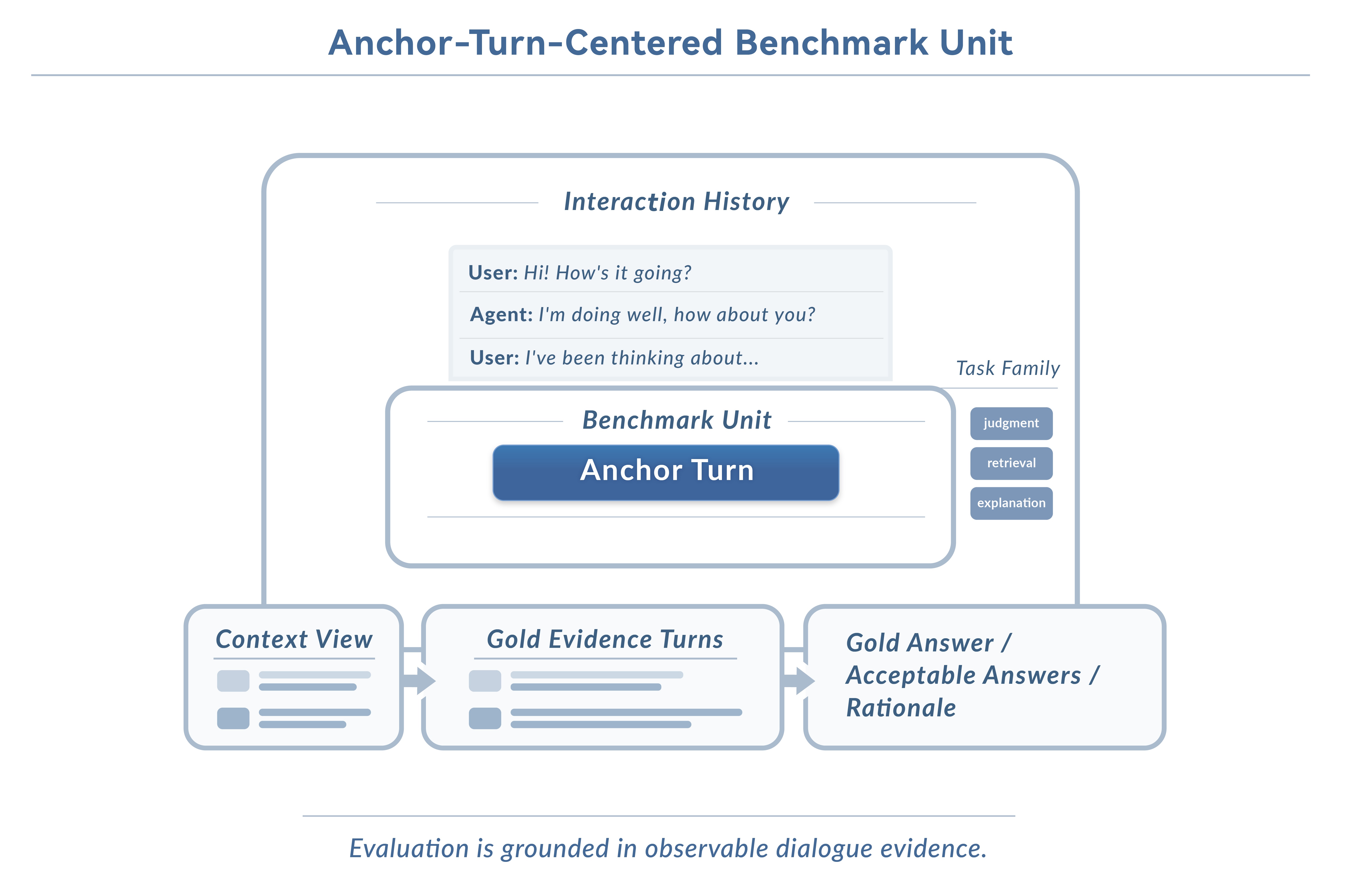}
\caption{Anchor-turn-centered benchmark unit schema in A-MBER. Each evaluation item is organized around a designated anchor turn together with its accessible interaction history, task-specific question, gold evidence, and expected output, so that present-state interpretation remains explicitly grounded in observable dialogue context.}
\label{fig:anchor_unit}
\end{figure}

A-MBER is organized around three primary task families. As Figure~\ref{fig:anchor_unit} makes clear, the model is not asked to summarize the whole interaction, but to interpret one anchor moment using the appropriate historical support: \emph{judgment}, \emph{retrieval}, and \emph{explanation}. Judgment tasks evaluate whether the model can infer the correct present affective or relational interpretation at the anchor turn. Retrieval tasks evaluate whether it can identify the historically relevant evidence turns rather than arrive at a plausible answer through superficial matching or chance. Explanation tasks evaluate whether it can justify the interpretation by linking the anchor turn to the relevant history. These task families are designed to work together so that the benchmark measures not only whether a model is correct, but whether it is correct for the right historical reasons.

The benchmark also includes control and robustness-oriented settings. Some items probe relatively local abilities, such as near-range factual understanding or more explicit affective interpretation, in order to separate basic conversational competence from genuine long-horizon memory use. Other items introduce modality-missing, modality-ambiguous, adversarial, or insufficient-evidence conditions. These settings test whether a model can remain calibrated when available cues are weak, incomplete, or potentially misleading. At the interface level, the public evaluation target is always grounded in observable dialogue evidence: each item is defined by a history window, an anchor turn, and a task-specific output space, while gold grounding is expressed through turn-level evidence rather than hidden generation notes.

\subsection{Scenario and Data Representation}

A-MBER is instantiated in a controlled interpersonal scenario rather than an unrestricted open-domain setting. In the present version, a teacher- or counselor-like agent interacts with a student across repeated sessions. This scenario provides stable role relations, meaningful event accumulation, and affective states whose interpretation often depends on history rather than on local cues alone. It also aligns naturally with the application logic of memory-centered interactive systems. At the same time, the framework itself is not restricted to this domain; the current scenario should be understood as a first controlled instantiation rather than the only valid setting.

Within this setting, the student side is treated as the primary affective target of evaluation, while the teacher or counselor side provides the supporting interaction context. This keeps the benchmark centered on whether a system can track and later interpret the user-side emotional trajectory across time. Interactions are organized as multi-session trajectories rather than as single extended dialogues. Sessions are linked through recurring topics, unresolved issues, support attempts, misunderstandings, partial repairs, and changing trust relations, allowing present turns to depend on accumulated history rather than on isolated local cues.

The benchmark is built around conversational interaction in which affective interpretation may depend on both textual content and delivery-related information. In the present version, this information is represented through dialogue text together with structured delivery descriptions that preserve affect-relevant vocal properties in a controlled benchmark format. This representation makes it possible to retain information that would otherwise be lost in plain transcript-only evaluation, while keeping the benchmark usable across model classes and evaluation settings.

The dataset uses layered structured representations rather than a single flat dialogue format. In the current implementation, each scenario is organized into linked layers including persona specification, long-horizon planning, conversation, annotation, question construction, and final benchmark units. These layers separate scenario design, observable interaction, turn-grounded supervision, task construction, and final benchmark packaging. This representation improves controllability during construction while preserving a clear distinction between internal construction support and public benchmark supervision.

\subsection{Primary Construction Pipeline}

In this study, A-MBER is primarily instantiated through a \emph{single-agent staged construction pipeline}. This route is used as the main production setting because it offers stronger controllability, easier schema alignment, and more stable large-scale benchmark production than freer interaction-based generation. The pipeline begins from a controlled production specification together with a persona pool, allowing diversity to arise from persona combinations and their associated interaction patterns within a shared scenario domain.

\begin{figure}[t]
\centering
\includegraphics[width=0.98\linewidth]{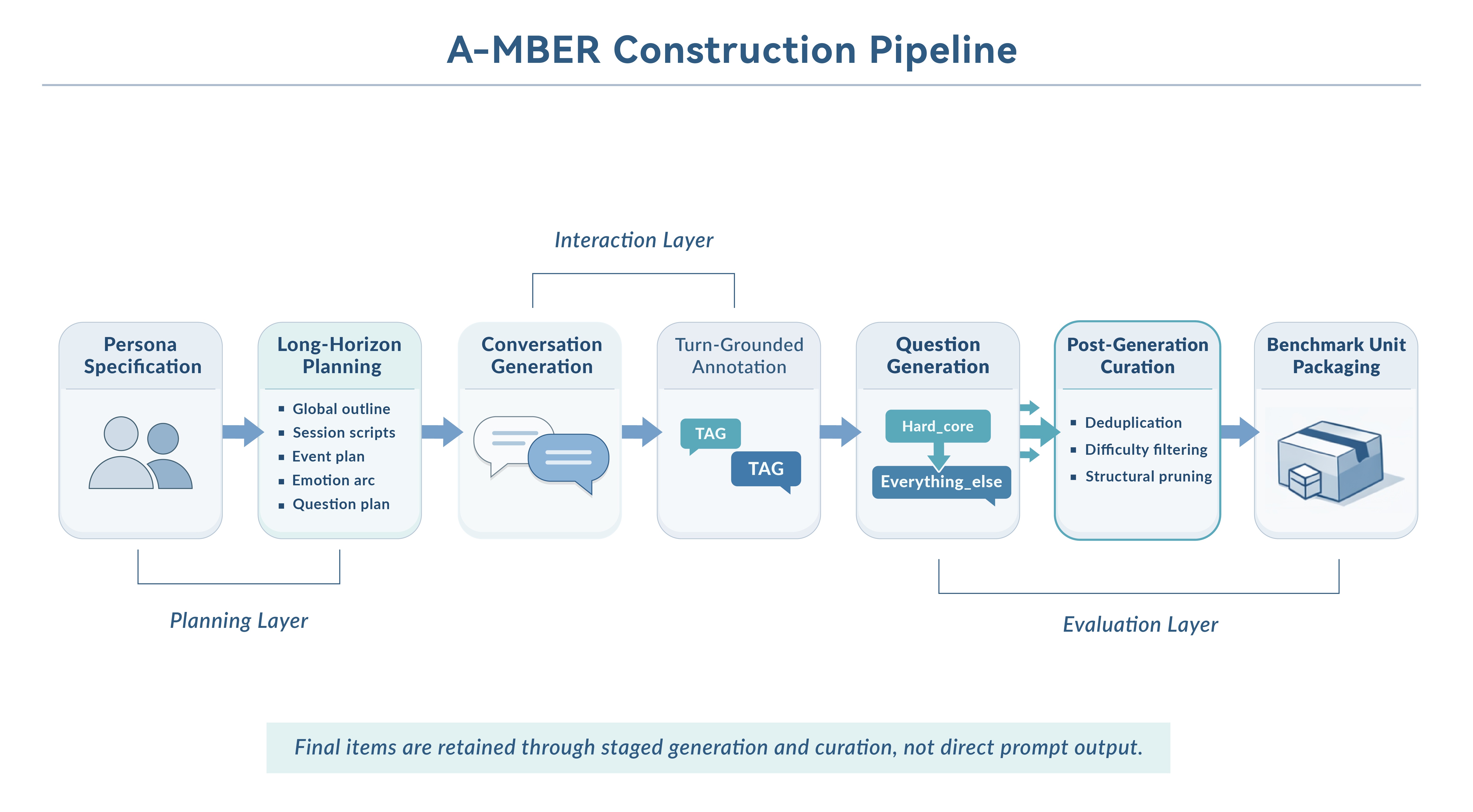}
\caption{Primary benchmark construction pipeline of A-MBER. The pipeline moves from persona and long-horizon planning to conversation generation, annotation, question construction, post-generation curation, and final benchmark-unit packaging, making the connection between scenario design, observable interaction, and evaluable supervision explicit.}
\label{fig:construction_pipeline}
\end{figure}

Figure~\ref{fig:construction_pipeline} provides the full construction view for this section and clarifies how staged intermediate representations are used to maintain controllability and evidence grounding throughout data generation. The first stage defines the long-horizon interaction backbone. For each scenario, the pipeline constructs a global outline, session scripts, an event plan, an emotion arc, and a question plan. Together, these components specify the cross-session structure of the interaction before surface dialogue is generated, including major events, affective developments, relation changes, and coverage requirements for downstream task construction. This planning-first design ensures that long-horizon dependency is built into the scenario itself rather than left to emerge incidentally from local dialogue generation.

On top of this backbone, the pipeline generates the observable conversation. The resulting interaction serves as the benchmark's primary evidence surface: downstream stages operate on dialogue turns rather than hidden planning notes alone. After conversation generation, the pipeline produces turn-grounded annotation, exposing the affective, relational, and evidential structure required for controlled benchmark construction. Gold evidence is tied to dialogue turn IDs, so that the benchmark remains aligned with what a model can actually observe.

Question construction is performed only after the conversation and annotation layers are in place. In A-MBER, this stage is not treated as a one-shot generation step. Question generation proceeds in two phases, with \texttt{hard\_core} items generated first and \texttt{everything\_else} generated afterwards. The resulting candidates are then curated through post-generation procedures such as deduplication, difficulty filtering, and structural pruning before they are retained as benchmark items. This part of the pipeline is important because it helps ensure that the final benchmark evaluates long-horizon affective reasoning under well-formed evidence conditions, rather than simply preserving all raw generated questions.

The final stage packages retained items into benchmark units centered on anchor turns. Each unit links the anchor turn, the relevant context view, the supporting evidence, the task specification, and the gold supervision into a benchmark-ready instance. This staged design serves three purposes. First, it improves controllability by separating long-horizon planning, dialogue realization, annotation, question construction, and final packaging into explicit layers. Second, it makes cross-session structure more reliable by planning critical events and affective trajectories before dialogue generation. Third, it preserves a clean distinction between internal construction support and public benchmark supervision. A-MBER should therefore be understood not as the product of one-shot generation, but as the result of structured benchmark construction.

\subsection{Alternative Interaction Realizations}

Although A-MBER is primarily instantiated through a single-agent staged construction pipeline, the overall framework is not tied to a single conversation-generation architecture. It also accommodates \emph{agent-to-agent} interaction as an auxiliary construction regime and can be extended to \emph{hardware-mediated} interaction settings in which persistent agents generate dialogue in a more embodied or deployment-oriented environment. These alternatives do not redefine the benchmark; they provide different ways of instantiating the conversation layer within the same broader construction logic.

In the present study, single-agent generation is preferred because it better supports controllability, schema alignment, turn-level grounding, and stable benchmark production. Multi-agent generation remains useful as a complementary regime because it may better preserve character differentiation, spontaneity, and emergent relational dynamics, but it is also more vulnerable to drift and evidence-structure instability under strong structural constraints. For the same reason, hardware-mediated interaction is better understood as a future realization path than as a separate benchmark definition. Across these settings, what changes is primarily the mechanism by which the conversation is produced; what remains stable is the broader framework of structured planning, turn-grounded annotation, question construction, and final packaging into historically grounded benchmark units.

\section{Benchmark Tasks and Evaluation Design}

\subsection{Task Families}

A-MBER is organized around three primary task families: \emph{judgment}, \emph{retrieval}, and \emph{explanation}. This design follows directly from the benchmark target of affective memory in long-horizon interaction. Rather than evaluating only whether a model reaches a plausible final interpretation, the benchmark also tests whether it can identify the relevant historical evidence and justify its answer in a history-grounded way. \emph{Judgment} tasks assess whether the model can infer the correct present affective or relational interpretation for a designated anchor turn. \emph{Retrieval} tasks assess whether it can identify the historically relevant supporting turns. \emph{Explanation} tasks assess whether it can justify the interpretation by linking the anchor turn to the appropriate history. Together, these task families distinguish surface correctness from historically grounded reasoning.

\begin{figure}[t]
\centering
\includegraphics[width=0.9\linewidth]{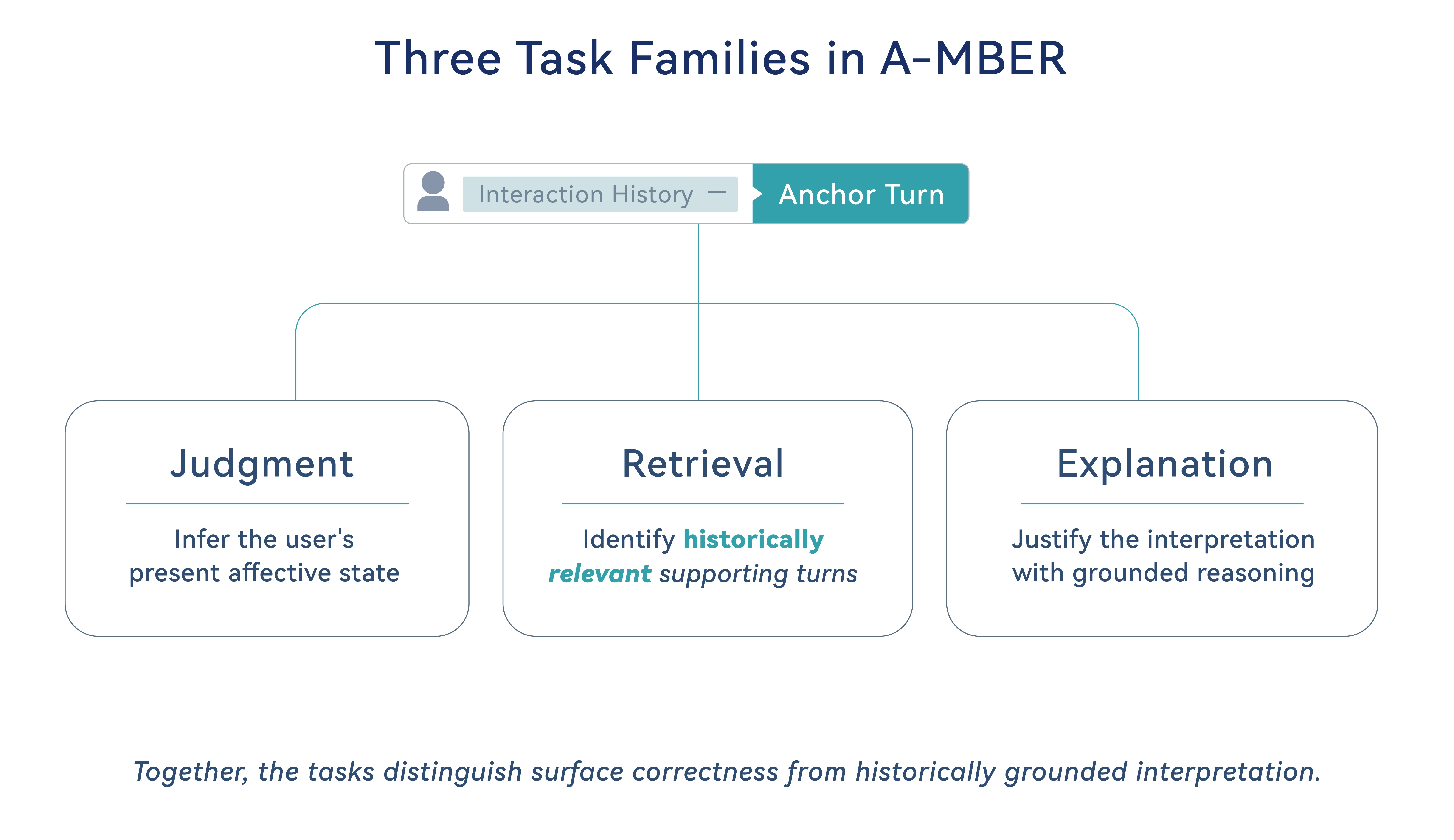}
\caption{Three primary task families in A-MBER. Judgment evaluates present-state interpretation at the anchor turn, retrieval evaluates whether the model can locate the historically relevant supporting turns, and explanation evaluates whether the interpretation can be justified in a grounded way.}
\label{fig:task_families}
\end{figure}

Beyond these primary task families, Figure~\ref{fig:task_families} summarizes the task decomposition used throughout the benchmark and makes clear that A-MBER evaluates both answer correctness and evidence-grounded reasoning. A-MBER also includes diagnostic and stress-test settings. The repository organizes the benchmark into a three-layer structure: a \emph{core} layer for the main evaluation distribution, a \emph{diagnostic} layer for interpretable sub-capability probes, and a \emph{stress-test} layer for robustness under degraded input conditions. Adversarial items, including pseudo-relevant-history and insufficient-evidence cases, are treated as a horizontal tag across layers rather than as a separate top-level family. This organization is intended to preserve a clear main evaluation target while still making it possible to diagnose why a model succeeds or fails.

\subsection{Benchmark Composition}

A-MBER is centered on long-horizon affective interpretation. Its main evaluation mass lies in items that require the model to use multi-session interaction history to infer the user's present affective state when the anchor turn is not fully interpretable from local wording alone. Factual and near-range items remain in the benchmark, but primarily as control tasks. Relation-state, trajectory-sensitive, and multi-hop items serve as diagnostic probes, while modality-sensitive items are used to test robustness when the locally available signal becomes weaker or less reliable.

\begin{figure}[t]
\centering
\includegraphics[width=0.94\linewidth]{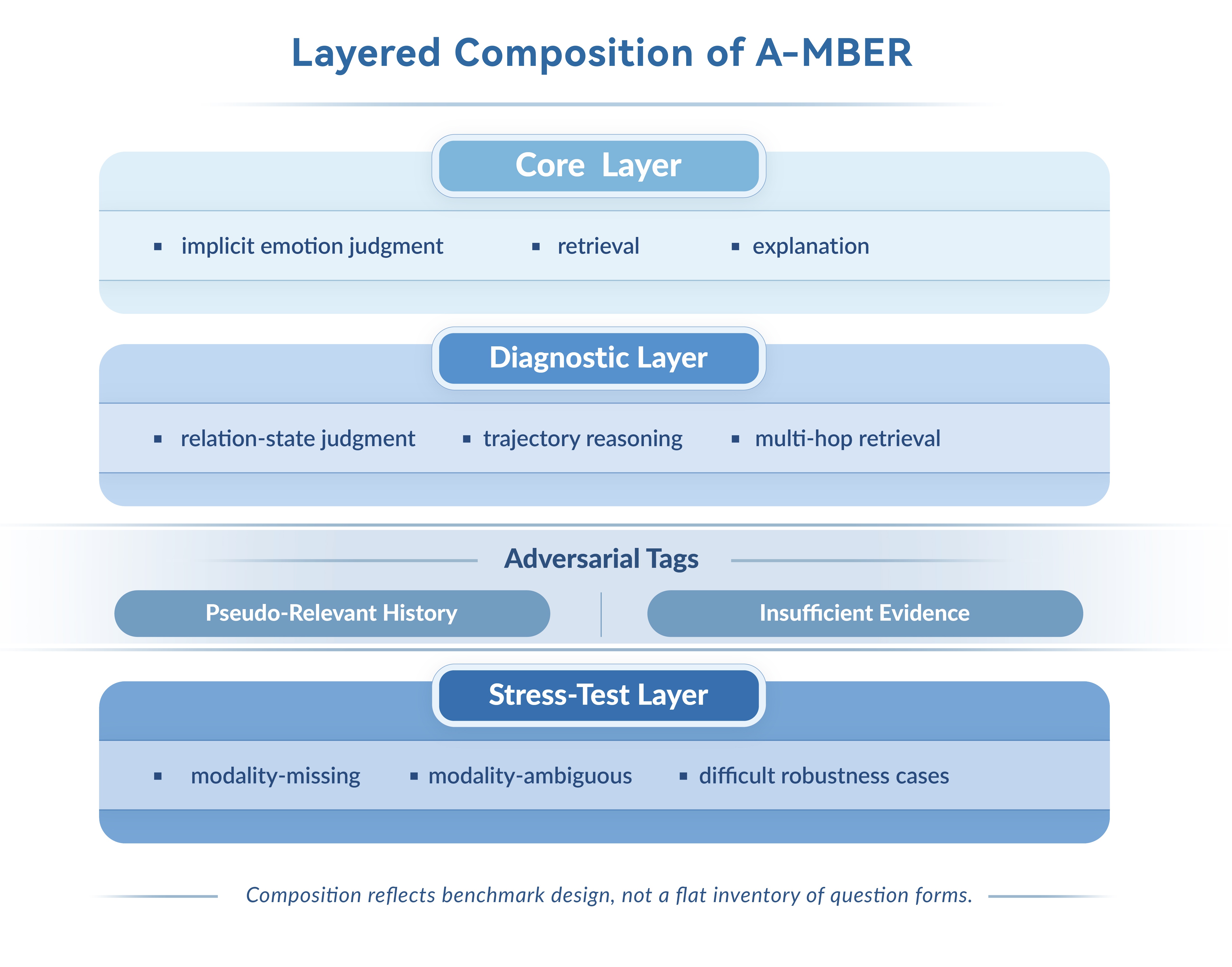}
\caption{Layered benchmark composition of A-MBER. The figure separates the core layer, diagnostic layer, and stress-test layer so that the main target of long-horizon affective interpretation remains distinct from auxiliary analyses of reasoning difficulty and robustness.}
\label{fig:layered_composition}
\end{figure}

This design is reflected in the benchmark's layered organization. As Figure~\ref{fig:layered_composition} makes explicit, the benchmark is centered on long-horizon affective-memory items, while diagnostic and stress-test subsets are included to interpret model behavior more precisely. The \emph{core} layer contains the main evaluation tasks, centered on implicit emotion judgment, retrieval, and explanation. The \emph{diagnostic} layer includes sub-capability probes such as relation-state judgment, trajectory-based reasoning, and multi-hop retrieval. The \emph{stress-test} layer introduces more difficult input conditions, including modality-missing and modality-ambiguous settings. Adversarial cases cut across these layers as horizontal tags. In this sense, benchmark composition is not merely a flat count of question forms; it is part of the benchmark's methodological design and helps separate the main target from the dimensions used to interpret model behavior. 

The final released composition is also shaped by post-generation curation. Since question generation is followed by deduplication, difficulty filtering, and structural pruning, the released benchmark is defined by the set of retained benchmark units rather than by every candidate produced during generation. The benchmark distribution is therefore a property of task design and post-generation quality control together, rather than of raw generation output alone. 

\subsection{Memory Levels and Reasoning Structure}

Not all benchmark items require the same degree of historical dependence. To make this explicit, each question is associated with a \emph{memory level} and a \emph{reasoning structure}. Memory levels indicate how strongly successful interpretation depends on prior interaction history. In the current design, these levels range from relatively local items to strongly history-dependent items, with the highest levels reserved for cases in which the anchor turn cannot be interpreted reliably without integrating longer-range emotional or relational context. 

Reasoning structure further specifies how interpretation depends on history. Some items are relatively direct and can be solved through local or single-hop reasoning. Others require \emph{multi-hop} or \emph{trajectory-based} reasoning, where multiple earlier events, relationship shifts, or recurrent patterns must be integrated in order to interpret the current moment. This distinction is particularly important for affective memory evaluation, because the challenge often lies not simply in retrieving one relevant fact, but in reconstructing a longer emotional and relational trajectory. 

Memory levels and reasoning structures function not only as descriptive metadata, but also as analysis dimensions. They support layered reporting beyond aggregate accuracy and make it possible to test whether access to longer interaction history is most helpful precisely on those items where historical reconstruction is most necessary.

\begin{figure}[t]
\centering
\includegraphics[width=0.94\linewidth]{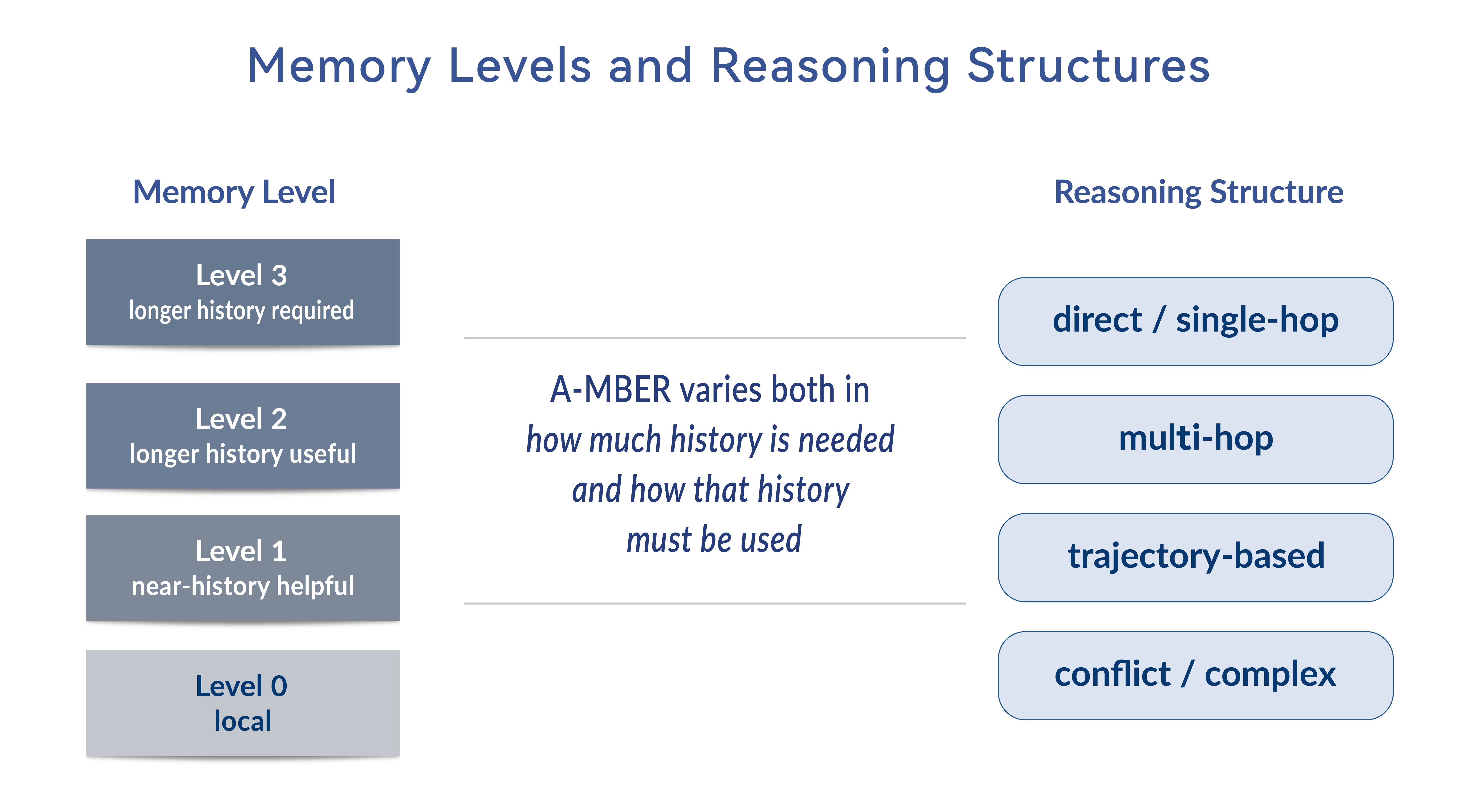}
\caption{Overview of memory levels and reasoning structures in A-MBER. The figure summarizes how items vary in historical dependency and in reasoning form, from relatively local or direct cases to strongly history-dependent, multi-hop, and trajectory-based interpretations.}
\label{fig:memory_reasoning}
\end{figure}

These distinctions define the main analysis dimensions used later in the experiments, and Figure~\ref{fig:memory_reasoning} helps explain why benchmark performance should be broken down by both memory dependency and reasoning complexity.

\subsection{Modality and Adversarial Conditions}

A-MBER includes modality-sensitive and adversarial conditions as robustness-oriented parts of the benchmark. In the released benchmark structure, modality-missing and modality-ambiguous items appear in the stress-test layer rather than serving as the primary organizing axis of the dataset. Their role is to test whether historically grounded interpretation remains stable when the locally available signal becomes weaker, less reliable, or partially unavailable. 

This design is consistent with the benchmark's broader focus on affective memory rather than on perceptual classification alone. The question is not simply whether a model can read the current signal correctly, but whether remembered interaction history can compensate when the present turn becomes harder to interpret from local evidence alone. For this reason, modality-sensitive items are most useful when read together with the benchmark's memory-level and reasoning-structure annotations, since they reveal whether longer-range evidence helps stabilize interpretation under degraded local conditions. 

A-MBER also includes adversarial settings. These include cases in which plausible but irrelevant history is present, as well as cases in which the available evidence is insufficient for a strong conclusion. Such items are designed to test whether a model can avoid over-weighting superficially plausible history or producing unjustified high-confidence interpretations. This is especially important for affective memory evaluation, where the interaction history may be subtle, incomplete, or underdetermined even when it appears emotionally suggestive on the surface. 

\subsection{Answer Evaluation and Scoring}

Because A-MBER contains multiple task formats, scoring is task-specific rather than uniform across all items. Judgment, retrieval, and explanation questions differ not only in output form but also in what counts as a successful answer, so the scoring framework follows the structure of the task while keeping supervision grounded in dialogue evidence. 

For \emph{judgment} items with predefined answer options, evaluation is based on exact match against the gold answer. For \emph{retrieval} items, evaluation is based on agreement between predicted and gold supporting turns. These task types are comparatively straightforward to score because they admit structured supervision tied directly to the interaction history. 

\emph{Explanation} items require a different evaluation principle. Many of them are intentionally open-form and are meant to test whether the model can connect the anchor turn to the appropriate prior interaction history in a coherent and historically grounded way. Exact string matching is therefore not sufficient. Instead, explanation scoring evaluates whether the response reaches the correct interpretation, uses the appropriate evidence, covers the essential reasoning steps, and avoids unsupported inference. In the released evaluation workflow, A-MBER provides built-in support for open-ended scoring through an \texttt{llm\_judge} evaluator, while option-style questions are scored through \texttt{exact\_match}. The scoring framework therefore combines direct automatic evaluation where the answer space is structured and judge-based semantic evaluation where historically grounded explanation is the main target. 

This issue becomes especially important for adversarial and insufficient-evidence items. In these settings, the benchmark is designed not only to reward correct interpretation, but also to measure whether a model remains calibrated when the available history is incomplete, weak, or misleading. A strong answer should therefore remain cautious when the interaction history does not warrant a strong conclusion. 

\begin{table}[t]
\centering
\small
\begin{tabular}{lll}
\toprule
\textbf{Task family} & \textbf{Answer format} & \textbf{Scoring} \\
\midrule
Judgment & Multiple choice / label & Exact match \\
Retrieval & Evidence turns & Set match / turn-level F1 \\
Explanation & Open-form text & LLM judge \\
Insufficient-evidence & Label / open-form & Exact match or calibration-sensitive judgment \\
\bottomrule
\end{tabular}
\caption{Task-specific scoring rules in A-MBER.}
\label{tab:task_scoring}
\end{table}

\subsection{Evaluation Protocol}

A-MBER evaluates present-state interpretation under different amounts of accessible conversational history. In the released evaluation workflow, runs are organized around explicit context policies rather than a single undifferentiated input setting. The repository currently supports two evaluation conditions: \texttt{session\_local}, in which the model sees only the current session up to the anchor turn, and \texttt{full\_history}, in which it sees the complete cross-session history up to that point. This contrast directly tests whether broader remembered interaction history improves affective interpretation relative to a more local view. 

Evaluation is conducted over benchmark units derived from the construction pipeline. Each dataset item is packaged from the benchmark-units layer and serves as a standardized evaluation instance for Langfuse-based experimentation. In this format, the question, anchor information, context view, expected output, and benchmark metadata are kept together, which makes it possible to compare models under matched benchmark instances while preserving the information needed for later analysis. 

At the reporting level, results should be presented both in aggregate form and through stratified analyses. Overall performance is useful, but it is not sufficient for understanding where memory matters most. For that reason, A-MBER is designed to support breakdowns by task family, content type, memory level, reasoning structure, and robustness condition. These analyses are necessary because the value of conversational memory is not expected to appear uniformly across all items. The benchmark is intended to show, in particular, whether access to longer interaction history is most helpful on long-horizon, affectively implicit, and evidence-sensitive questions. 

The same logic applies to modality-sensitive items. In A-MBER, these items function as robustness-oriented settings within the broader affective-memory benchmark rather than as a separate benchmark target. Their role is to test whether historically grounded interpretation remains stable when the local signal becomes weaker, less reliable, or partially unavailable, and thus whether longer-range evidence can compensate when the present turn becomes less directly interpretable.

\section{Experiments}

\subsection{Experimental Goals}

The experiments are designed to answer two related questions. First, does access to memory improve performance on affective-memory tasks relative to settings in which the model relies only on limited local context? Second, if so, where does this advantage come from: broader access to interaction history, better selection of relevant evidence, or stronger use of that history for long-horizon affective interpretation under incomplete, ambiguous, or potentially misleading local conditions?

A-MBER is not intended merely as a benchmark on which one system obtains a higher aggregate score than another. Its purpose is to test whether memory becomes especially useful on the kinds of items for which present-state interpretation depends on remembered history rather than local wording alone. For this reason, the experiments are organized not only around overall performance, but also around several benchmark-specific dimensions, including content type, memory level, reasoning structure, and robustness condition. These analyses make it possible to ask not only whether memory helps, but where it helps most and what kind of historical reasoning that improvement reflects.

\subsection{Compared Systems}

We compare five system configurations.

\paragraph{No-Memory Baseline.}
This baseline receives only the anchor turn or minimal local context, without access to broader cross-session history. It serves as the lower-bound comparison for testing how much present-state interpretation can be recovered from local evidence alone.

\paragraph{Long-Context Baseline.}
This baseline is given a larger raw window of interaction history but no explicit retrieval or structured memory support. It tests whether simply exposing more history is sufficient for strong affective interpretation.

\paragraph{Retrieved-Memory Baseline.}
This baseline answers using the current input together with a retrieved subset of historical context. It isolates the effect of explicit retrieval and allows comparison between raw long context and selected supporting history.

\paragraph{Structured Memory System.}
This configuration corresponds to a system with explicit memory organization, such as the Red Bear AI memory system. The aim is to test whether structured memory helps not only with access to history, but also with its effective use for affect-sensitive interpretation.

\paragraph{Gold-Evidence Condition.}
This optional condition directly provides gold supporting history derived from benchmark annotation. It is not intended as a realistic baseline, but as an analysis tool for estimating how much of the remaining difficulty comes from evidence selection as opposed to downstream interpretation.

Together, these settings define a comparison path from minimal local context, to broader raw context, to retrieved evidence, to structured memory, and finally to gold-evidence access. This progression makes it possible to separate several distinct sources of performance gain that would otherwise be conflated in a single overall system comparison.

\subsection{Experimental Conditions}

To ensure fair comparison, all systems are evaluated under explicitly defined input conditions. At the context level, the main comparisons include local-only settings, full-history settings, retrieved-history settings, structured-memory settings, and gold-evidence conditions. Keeping these conditions explicit is important, since different systems may otherwise differ not only in memory quality but also in the amount and form of information made available to them.

All systems are evaluated on matched benchmark units with the same anchor turns, gold evidence, and task definitions. This makes it possible to attribute performance differences to the memory setting itself rather than to uncontrolled variation in the underlying evaluation data. The same benchmark structure is used across judgment, retrieval, and explanation tasks, so that differences across systems can be interpreted within a single unified benchmark framework rather than across unrelated evaluation setups.

In addition, A-MBER includes robustness-oriented subsets such as modality-missing, modality-ambiguous, and adversarial items. These are not treated as separate benchmarks, but as controlled conditions within the same evaluation framework. Their purpose is to test whether access to longer-range evidence remains useful when the locally available signal is weak, incomplete, or potentially misleading. This is especially important for affective-memory evaluation, since many realistic interaction failures do not arise from the complete absence of information, but from situations in which local evidence is suggestive yet insufficient.

\begin{figure}[t]
\centering
\includegraphics[width=0.98\linewidth]{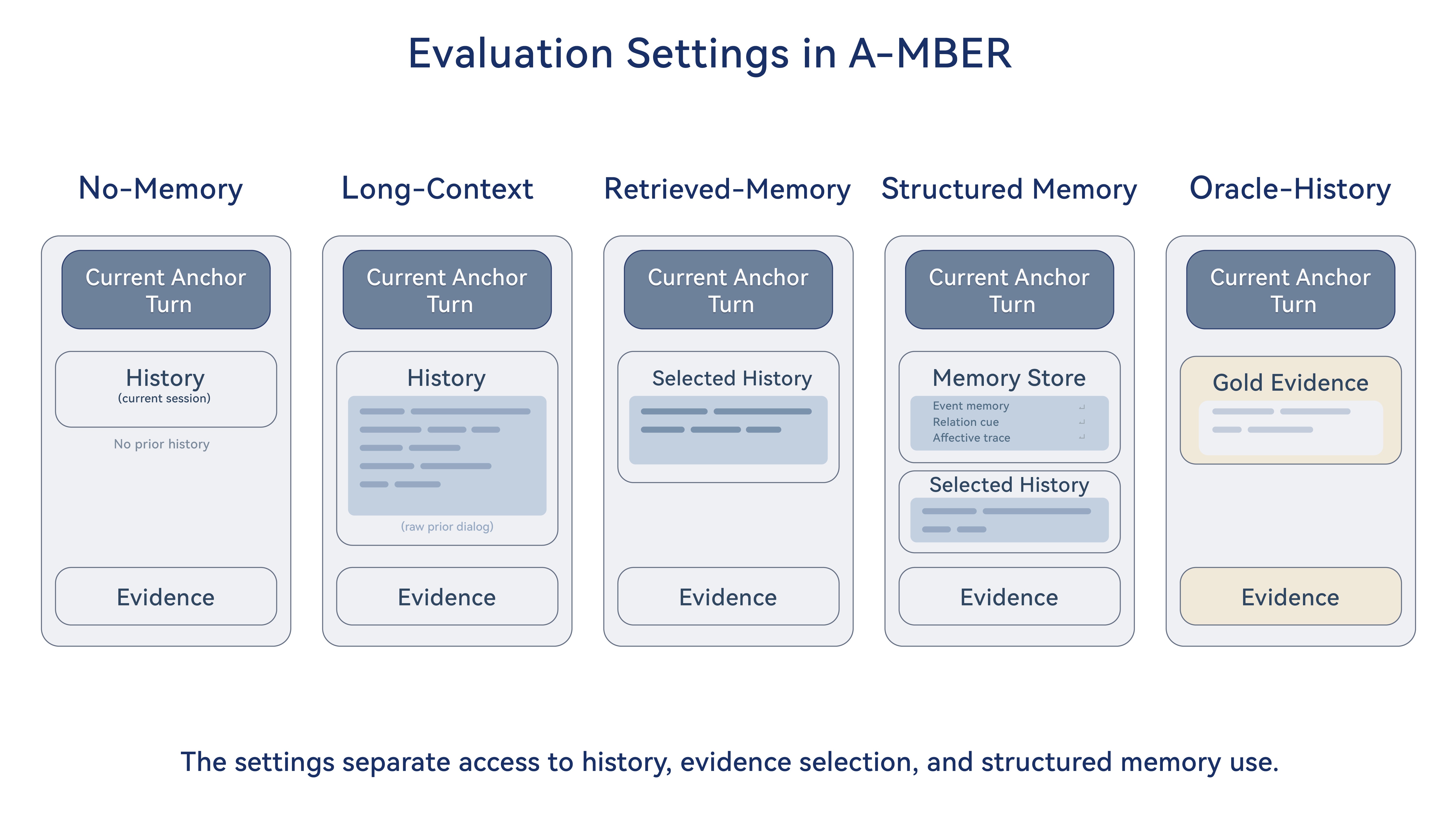}
\caption{Evaluation settings used for system comparison in A-MBER. The figure contrasts no-memory, long-context, retrieved-memory, structured-memory, and gold-evidence conditions, making explicit how each system is given different forms of historical access while keeping the benchmark units themselves fixed.}
\label{fig:evaluation_settings}
\end{figure}

Figure~\ref{fig:evaluation_settings} visualizes the comparison path used in the experiments and clarifies that the evaluation changes the form and quality of accessible history rather than the underlying benchmark items.


\subsection{Main Results}

Table~\ref{tab:main_results} reports the main comparison across system configurations. This table is the core experiment of the paper. The most important pattern is whether performance improves systematically as access to history becomes broader and more structured, and whether structured memory provides an advantage beyond raw long-context exposure or retrieval alone.

\begin{table}[t]
\centering
\small
\begin{tabular}{lccc}
\toprule
\textbf{System} & \textbf{Judgment} & \textbf{Retrieval} & \textbf{Explanation} \\
\midrule
No-Memory Baseline & 0.34 & 0.29 & 0.31 \\
Long-Context Baseline & 0.47 & 0.41 & 0.44 \\
Retrieved-Memory Baseline & 0.58 & 0.54 & 0.53 \\
Red Bear AI Memory & \textbf{0.69} & \textbf{0.66} & \textbf{0.65} \\
Gold-Evidence & 0.81 & 0.79 & 0.77 \\
\bottomrule
\end{tabular}
\caption{Results across compared systems on A-MBER.}
\label{tab:main_results}
\end{table}

Table~\ref{tab:main_results} shows a clear progression across memory settings. The no-memory baseline performs worst across all three task families, indicating that a substantial portion of A-MBER cannot be solved reliably from the anchor turn or minimal local context alone. Providing broader raw history already improves performance consistently, but the benchmark remains challenging even when more history is made available in raw form.

A second improvement appears when the system moves from raw long context to retrieved history. This pattern indicates that access to more history is not the whole story; selecting the historically relevant portion of that history also matters, but selection alone still leaves substantial room for improvement. In particular, the retrieval gains support the idea that A-MBER rewards not only remembering more context, but identifying the evidence that is actually relevant to the anchor turn.

The strongest non-gold-evidence performance is obtained by the structured memory system. This is a central result for the benchmark. It suggests that A-MBER does not merely reward larger context windows or naive retrieval, but is sensitive to whether remembered history is organized and used effectively for affect-sensitive interpretation. The remaining gap to the gold-evidence condition further suggests that there is still room for improvement both in evidence access and in downstream use of the selected history. At the same time, the fact that gold-evidence performance remains clearly below ceiling indicates that the benchmark retains substantial interpretation difficulty even when gold supporting evidence is available.

\subsection{Results by Content Type}

A-MBER is intentionally centered on long-horizon affective interpretation, but it also includes factual, local, and robustness-oriented subsets. Breaking results down by content type makes it possible to test whether memory helps most on the benchmark's main target rather than uniformly across all item types.

\begin{table}[t]
\centering
\small
\begin{tabular}{lccccc}
\toprule
\textbf{System} & \textbf{Implicit} & \textbf{Explicit} & \textbf{Instant} & \textbf{Long Fact} & \textbf{Near Fact} \\
\midrule
No-Memory & 0.18 & 0.31 & 0.45 & 0.27 & 0.49 \\
Long-Context & 0.34 & 0.46 & 0.57 & 0.41 & 0.60 \\
Retrieved-Memory & 0.49 & 0.58 & 0.66 & 0.54 & 0.68 \\
Red Bear AI Memory & \textbf{0.65} & \textbf{0.69} & \textbf{0.74} & \textbf{0.67} & \textbf{0.76} \\
Gold-Evidence & 0.79 & 0.81 & 0.84 & 0.78 & 0.85 \\
\bottomrule
\end{tabular}
\caption{Results by content type. ``Implicit'' denotes long-range implicit affect; ``Explicit'' denotes long-range explicit or semi-explicit affect.}
\label{tab:content_results}
\end{table}

The content-type breakdown in Table~\ref{tab:content_results} is closely aligned with the intended target of A-MBER. The largest gains from memory appear on long-range implicit affect items. This is precisely the pattern one would want from a benchmark centered on affective memory: the strongest memory benefit appears in cases where the meaning of the anchor turn is underdetermined unless the system reconstructs the relevant emotional history. The especially low local-only performance on this subset is therefore not merely a sign of difficulty, but evidence that the benchmark is successfully concentrating its hardest cases where remembered affective history matters most.

By contrast, the gains are smaller on instant-emotion and near-range factual items. These differences are still meaningful, but they are less diagnostic of long-horizon affective memory than the implicit subset. Long-range factual items occupy an intermediate position: they benefit from memory, but less sharply than implicit affect items. Taken together, this pattern supports the claim that A-MBER is centered on historically grounded affective interpretation rather than generic long-term question answering.

\subsection{Results by Memory Level}

Memory-level analysis is central to A-MBER because the benchmark is intentionally skewed toward stronger history dependence. If memory support is genuinely useful, the advantage of history-aware systems should grow as the required historical dependence becomes stronger.

\begin{table}[t]
\centering
\small
\begin{tabular}{lcccc}
\toprule
\textbf{System} & \textbf{Level 0} & \textbf{Level 1} & \textbf{Level 2} & \textbf{Level 3} \\
\midrule
No-Memory & 0.52 & 0.39 & 0.28 & 0.16 \\
Long-Context & 0.62 & 0.50 & 0.40 & 0.30 \\
Retrieved-Memory & 0.70 & 0.61 & 0.53 & 0.43 \\
Red Bear AI Memory & \textbf{0.78} & \textbf{0.71} & \textbf{0.64} & \textbf{0.58} \\
Gold-Evidence & 0.84 & 0.80 & 0.76 & 0.72 \\
\bottomrule
\end{tabular}
\caption{Results by memory dependency level.}
\label{tab:memory_level_results}
\end{table}

The results by memory level in Table~\ref{tab:memory_level_results} provide one of the clearest validations of the benchmark design. As expected, performance differences across systems are relatively smaller on Level~0 and Level~1 items, where local or near-term context may already be sufficient. Even so, memory still produces a measurable benefit at these lower levels. At the same time, the benchmark is not trivial even there, which helps prevent the evaluation from collapsing into a purely easy-local versus hard-history dichotomy.

The picture changes substantially at higher levels of historical dependence. The largest gains appear on the most history-dependent items, especially at Level~3. This is an especially important result because Level~3 is intended to capture precisely those cases in which present interpretation depends on substantial cross-session reconstruction. The fact that the largest gains appear there strongly supports the benchmark's internal logic. It suggests that A-MBER's memory-level annotations are not merely descriptive metadata, but correspond to meaningful differences in the degree to which remembered history is required for successful interpretation.

\subsection{Results by Reasoning Structure and Robustness Condition}

We further analyze results by reasoning structure and robustness condition. Reasoning-structure analysis tests whether memory is most beneficial on items that require integrating multiple parts of the interaction history rather than on more direct or single-hop cases. Robustness analysis tests whether remembered history remains useful when the locally available signal is degraded or potentially misleading.

\begin{table}[t]
\centering
\small
\begin{tabular}{lcccc}
\toprule
\textbf{System} & \textbf{Direct / Single-hop} & \textbf{Multi-hop} & \textbf{Trajectory} & \textbf{Conflict / Complex} \\
\midrule
No-Memory & 0.49 & 0.27 & 0.21 & 0.24 \\
Long-Context & 0.60 & 0.39 & 0.33 & 0.36 \\
Retrieved-Memory & 0.69 & 0.53 & 0.47 & 0.50 \\
Red Bear AI Memory & \textbf{0.77} & \textbf{0.68} & \textbf{0.63} & \textbf{0.60} \\
Gold-Evidence & 0.84 & 0.79 & 0.75 & 0.72 \\
\bottomrule
\end{tabular}
\caption{Results by reasoning structure.}
\label{tab:reasoning_results}
\end{table}

\begin{table}[t]
\centering
\small
\begin{tabular}{lcccc}
\toprule
\textbf{System} & \textbf{Standard} & \textbf{Modality-Missing} & \textbf{Modality-Ambiguous} & \textbf{Adversarial} \\
\midrule
No-Memory & 0.39 & 0.30 & 0.28 & 0.22 \\
Long-Context & 0.52 & 0.41 & 0.38 & 0.32 \\
Retrieved-Memory & 0.62 & 0.52 & 0.49 & 0.43 \\
Red Bear AI Memory & \textbf{0.73} & \textbf{0.64} & \textbf{0.60} & \textbf{0.54} \\
Gold-Evidence & 0.82 & 0.75 & 0.72 & 0.66 \\
\bottomrule
\end{tabular}
\caption{Results by robustness condition.}
\label{tab:robustness_results}
\end{table}

Table~\ref{tab:reasoning_results} shows that the benefit of memory is not evenly distributed across reasoning structures. On direct or single-hop items, all systems perform comparatively better, and the gap between the no-memory baseline and structured memory remains more limited than on the more history-dependent subsets. By contrast, the differences become much larger on multi-hop and trajectory-based questions. These are among the largest gains in the chapter and are highly consistent with the benchmark's intended difficulty profile.

This pattern matters because it suggests that A-MBER is especially sensitive to cases in which interpretation depends on integrating multiple pieces of history rather than retrieving one isolated fact. In particular, trajectory-based items appear to function as a strong test of affective memory, since they require the model to reconstruct how the user's emotional state develops across sessions rather than simply identify a single relevant event. The conflict or complex subset also benefits from stronger memory settings, although somewhat less sharply, which is plausible given that these items combine evidence integration with more locally ambiguous interpretation.

The robustness results in Table~\ref{tab:robustness_results} show a similar pattern. All systems perform best under standard conditions and degrade when the local signal becomes weaker or less reliable. However, the structured memory system declines less sharply than the weaker baselines, especially under modality-missing, modality-ambiguous, and adversarial settings. This supports one of the central hypotheses of the paper: remembered history can compensate when present-time evidence is incomplete or uncertain. In other words, the value of memory in A-MBER is not only that it improves average performance, but that it stabilizes interpretation when local cues become less dependable.

\subsection{Results on Adversarial and Insufficient-Evidence Items}

A-MBER also evaluates whether systems remain calibrated under adversarial and underdetermined conditions. These subsets test not only whether a system can reach correct answers, but whether it can avoid over-interpreting plausible but irrelevant history or forcing a strong conclusion when the available evidence is insufficient.

\begin{table}[t]
\centering
\small
\begin{tabular}{lccc}
\toprule
\textbf{System} & \textbf{Insufficient Evidence} & \textbf{Pseudo-Relevant History} & \textbf{Other Adversarial} \\
\midrule
No-Memory & 0.19 & 0.24 & 0.28 \\
Long-Context & 0.31 & 0.38 & 0.42 \\
Retrieved-Memory & 0.43 & 0.51 & 0.55 \\
Red Bear AI Memory & \textbf{0.56} & \textbf{0.63} & \textbf{0.66} \\
Gold-Evidence & 0.68 & 0.75 & 0.77 \\
\bottomrule
\end{tabular}
\caption{Results on adversarial and insufficient-evidence subsets.}
\label{tab:adversarial_results}
\end{table}

The adversarial breakdown in Table~\ref{tab:adversarial_results} highlights another important aspect of the benchmark: calibration. The lowest scores appear on insufficient-evidence items, and even the stronger memory settings remain clearly below ceiling. This is a useful pattern. It suggests that these items remain genuinely difficult even when stronger memory support is available, which is exactly what one would expect from cases designed to penalize unwarranted certainty. This is a desirable property rather than a weakness of the benchmark, since insufficient-evidence cases are intended to remain difficult even under stronger history access.

Pseudo-relevant-history items show a similar but slightly less severe trend. Performance improves steadily across memory settings, but the task remains challenging because the system must avoid over-weighting plausible but ultimately irrelevant historical evidence. The fact that structured memory outperforms both long context and retrieved memory on these subsets suggests that the advantage of memory is not limited to accessing more evidence. It also includes using that evidence more cautiously and selectively. This is important for the overall argument of the paper, since affective-memory systems should not only recover the past more effectively, but also avoid turning remembered history into a source of overconfident error.

\subsection{Auxiliary Construction Analysis}

In addition to the main system comparison, we also examine benchmark construction itself. The purpose of this analysis is not to claim that one construction regime is universally better, but to clarify the trade-off between controllability and realism in long-horizon benchmark production.

In the current study, single-agent staged generation remains the primary construction route because it provides stronger controllability, better schema stability, and easier alignment between planning, annotation, and final benchmark units. Multi-agent generation is retained as a complementary regime with potential advantages in character differentiation, spontaneity, and relational naturalness. The purpose of this auxiliary analysis is therefore not to rank one regime categorically above the other, but to clarify the trade-off between production stability and interactional realism.

\begin{table}[t]
\centering
\small
\begin{tabular}{lcc}
\toprule
\textbf{Dimension} & \textbf{Single-Agent} & \textbf{Multi-Agent} \\
\midrule
Schema stability & High & Moderate \\
Long-horizon consistency & High & Moderate \\
Character differentiation & Moderate & High \\
Interaction spontaneity & Moderate & High \\
Question derivation reliability & High & Moderate \\
Overall production suitability & Primary & Auxiliary \\
\bottomrule
\end{tabular}
\caption{Comparison of single-agent and multi-agent construction regimes.}
\label{tab:construction_comparison}
\end{table}

The auxiliary construction analysis is included to clarify how benchmark production choices affect the final evaluation resource. In the present study, single-agent staged generation remains the primary route because it provides stronger controllability, more stable schema alignment, and cleaner linkage between planning, annotation, and final benchmark units. Multi-agent generation remains promising as a complementary regime, particularly for character differentiation and interaction spontaneity, but its advantages are currently more methodological than central to the benchmark's main claim.

For this reason, the comparison in Table~\ref{tab:construction_comparison} should be interpreted as a construction-side analysis rather than a headline experimental result. Its main purpose is to explain why the present version of A-MBER relies on a staged single-agent pipeline while still leaving room for future comparison under richer interaction regimes.

\section{Discussion and Future Extensions}

\subsection{What Current Benchmarks Still Miss}

A-MBER is motivated by a gap that remains only partially addressed by existing resources. Long-term conversational memory benchmarks have made important progress in evaluating multi-session reasoning, temporal tracking, and long-range information retention, while emotion and multimodal conversation datasets have provided valuable resources for utterance-level emotion recognition, empathy, and explanation. Yet these two lines of work still leave limited support for evaluating whether a system can use emotionally meaningful history to interpret a user's current state across sessions.

This gap matters because many real interaction failures do not arise from forgetting an isolated fact. They arise when a system fails to recognize how prior disappointments, recurring triggers, changing support preferences, or accumulated relationship dynamics alter the meaning of the present moment. A system may therefore perform well on factual memory benchmarks and still behave in affectively inappropriate ways if it cannot reconstruct the user's emotional trajectory in a historically grounded manner. A-MBER is intended to make this missing capability measurable.

For the same reason, A-MBER should not be understood as competing with existing long-context or long-memory benchmarks on the basis of raw interaction length alone. In our setting, long-horizon difficulty is defined not only by turn count or text length, but also by the density of event dependencies, affect-relevant history, and cross-session interpretive burden. Even when an interaction is shorter than those used in some very-long-context benchmarks, it may still impose substantial long-range reasoning demands if present interpretation depends on a tightly coupled emotional and relational history.

The current experimental results further clarify what this benchmark target entails. In particular, the low local-only baseline is not an accidental by-product of question difficulty, but a consequence of concentrating the benchmark's main mass on cases where present interpretation is not reliably recoverable from the anchor turn alone. At the same time, the fact that gold-evidence performance remains clearly below ceiling indicates that A-MBER does not reduce to a pure evidence-access benchmark. Even when gold supporting evidence is provided, historically grounded affective interpretation retains substantial difficulty beyond evidence selection alone.

\subsection{Why Affective Memory Matters for Memory-Centered Systems}

Affective memory matters for memory-centered systems because the value of long-term interaction depends on more than persistent storage. A system may record extensive user history, retrieve superficially relevant episodes, or maintain large context windows, yet still fail to respond in a way that is affectively appropriate. What matters is not only whether the system remembers more, but whether it uses remembered history to interpret the user more appropriately over time.

This is precisely the perspective that A-MBER is designed to evaluate. Its value lies in separating different possible sources of system advantage. A model may perform well because it sees more context, because it retrieves better evidence, or because it organizes and uses remembered history more effectively. By including no-memory, long-context, retrieved-memory, structured-memory, and gold-evidence comparison settings, the benchmark makes it possible to ask not only whether a memory system performs better overall, but where that advantage comes from. This is particularly relevant for memory-centered systems such as Memory Bear, whose intended strength lies not merely in exposing more historical content, but in maintaining and deploying memory in a structured way for affect-sensitive downstream interpretation.

The present results are consistent with this perspective. Performance improves from local-only access to broader history, then again from raw history to selected history, and again from retrieval to structured memory. This progression supports a stronger claim than “more context helps.” It suggests that affective memory in long-horizon interaction depends not only on seeing more of the past, but on selecting, organizing, and using remembered history in a way that is relevant to the current moment.

\subsection{Limitations of the Current Benchmark}

The current benchmark has several limitations. First, although the construction pipeline is structured and grounded in explicit intermediate representations, it still depends on generated interaction rather than naturally occurring human conversation. As a result, some aspects of real interpersonal dynamics remain simplified, especially timing irregularities, conversational derailments, subtle inconsistencies, and the broader messiness of long-term interaction. More generally, synthetic construction can make event structure, evidence chains, and affective dependencies cleaner than they would typically be in realistic dialogue, even when the resulting data remain useful for controlled evaluation.

Second, the benchmark is currently instantiated in a single controlled scenario with a relatively narrow relational structure, namely a teacher--student or counselor--student style interaction. This is a reasonable starting point because it provides stable role relations, emotionally meaningful trajectories, and clear alignment with long-term support settings. At the same time, it limits immediate claims of domain generality. The framework is intended to transfer across broader scenarios, but this remains to be demonstrated for interaction settings with less stable role boundaries, such as companionship, peer interaction, customer support, or care-oriented dialogue.

Third, the benchmark still reflects a constrained task formalization. The anchor-turn interface is useful because it makes present-state interpretation measurable in a controlled way, but it remains a benchmark interface rather than a complete model of real deployment. In actual interaction, users do not present their internal state as an item with a predefined evidence boundary. A-MBER therefore captures an important and previously under-evaluated capability, but it does so through a task-oriented slice of long-term interaction rather than through an exhaustive simulation of real-world use.

Fourth, although the benchmark includes open-form explanation tasks, their evaluation still depends in part on rubric-guided semantic judgment. Closed-form items admit relatively direct scoring, but explanation items inevitably retain some dependence on evaluator design, acceptable-answer coverage, and judge stability. This does not undermine the usefulness of such evaluation, but it does mean that fine-grained score differences on open-form items should be interpreted with appropriate caution.

Fifth, the current robustness conditions remain limited relative to the full range of ambiguity encountered in real interaction. Modality-missing, modality-ambiguous, and insufficient-evidence settings are useful first steps, but they still cover only part of the broader space of prolonged contradiction, partial observability, and complex multimodal inconsistency. A related practical challenge is that, because the benchmark is intentionally concentrated on historically dependent and affectively implicit cases, local-only performance may be quite low even for strong models. This is partly a design goal, but it also increases the importance of careful answer calibration, evaluator stability, and subset-level analysis when interpreting overall scores.

Similarly, while the benchmark framework can accommodate both single-agent and multi-agent construction, the present study does not yet provide a definitive comparison between these regimes under substantially longer trajectories. Single-agent staged generation is used as the primary route because it provides stronger controllability and more stable alignment, whereas the longer-horizon advantages of multi-agent generation remain to be established more fully.

Finally, our notion of long-horizon difficulty is intentionally task-oriented. We argue that long-range difficulty should be defined not only by raw length, but also by event density, affect-relevant dependency, and cross-session interpretive burden. We believe this is the appropriate framing for affective memory, but it is not the only possible definition of long-range difficulty. A-MBER should therefore be read as one principled perspective on long-horizon affective reasoning rather than as a single definitive standard for all long-context evaluation.

\subsection{Future Extensions}

Several extensions would strengthen the benchmark in future versions. The first is broader scenario and relation coverage. Extending A-MBER beyond the current teacher--student style setting would make it possible to test whether the same benchmark logic holds across different affective and relational structures, including companionship, peer support, customer interaction, and longer-term care settings.

A second extension concerns richer perceptual evidence. In the current benchmark, affective interpretation is grounded primarily in dialogue text together with structured delivery-related information. Future versions could incorporate stronger audio-grounded settings, visual cues, facial expression, gaze, posture, or broader interaction context. Such extensions would allow more realistic evaluation of multimodal affective memory and would make it possible to study cross-modal inconsistency more directly. At the same time, this direction should be approached carefully, since richer perceptual signals also increase annotation complexity and make controlled evaluation more difficult.

A third extension is stronger evaluation-side validation. In particular, future work should examine the robustness of open-form scoring more directly, for example through multi-judge comparison, human validation, or stronger agreement analysis between automatic and judge-based evaluation. This would strengthen the benchmark as explanation tasks become more central.

A fourth extension concerns data realism. Although synthetic construction provides controllability and coverage, future versions of the benchmark could benefit from human-written, human-revised, or partially human-collected interaction data. Such settings would help test whether the benchmark's core notion of affective memory remains stable under noisier and less cleanly structured dialogue. More embodied or device-mediated interaction may also become relevant in this broader context, not as the center of the benchmark definition, but as one possible route toward more realistic long-term interaction traces.

\subsection{Broader Outlook}

More broadly, we view A-MBER as a first step toward a richer evaluation ecosystem for affective memory in long-horizon interaction. The current version establishes a controlled benchmark target, a structured construction framework, and a set of evaluation dimensions that make historically grounded affective interpretation measurable. Future work can extend this ecosystem along several axes: broader scenario coverage, richer multimodal inputs, stronger evaluator validation, more realistic data collection, and deeper comparison across different memory architectures and construction regimes.

In this sense, A-MBER is intended not only as a static dataset, but as infrastructure for studying how memory supports affective understanding over time. Its main contribution is to make a previously under-measured capability visible: not simply whether a system remembers the past, but whether it can use remembered history to understand what the user is feeling now.

\section{Conclusion}

This paper introduces A-MBER, an \emph{Affective Memory Benchmark for Emotion Recognition}, to evaluate a capability that remains insufficiently captured by existing long-term memory benchmarks and existing emotion or multimodal conversation datasets: using remembered multi-session interaction history to support historically grounded interpretation of a user's present affective state. In this sense, A-MBER is designed to evaluate memory not merely as storage, but as an active resource for affective understanding.

To make this capability measurable, we propose a structured benchmark construction framework based on staged generation and explicit intermediate representations. In the current study, this framework is primarily instantiated through a single-agent construction pipeline for controllability and schema stability, while remaining compatible with alternative interaction realizations under the same benchmark logic. The resulting benchmark supports judgment, retrieval, and explanation tasks, together with layered analysis by content type, memory level, reasoning structure, and robustness condition.

Our experiments show that A-MBER is not easily solvable from local context alone, and that performance improves consistently as systems are given broader, more selective, and more structured access to history. The strongest gains appear precisely where the benchmark is intended to be most demanding: long-range implicit affect, highly history-dependent items, trajectory-based reasoning, and adversarial conditions. At the same time, the fact that gold-evidence performance also remains below ceiling suggests that the benchmark preserves substantial interpretation difficulty even when gold supporting evidence is available.

More broadly, A-MBER is intended to support the evaluation of memory-centered interactive systems, including systems such as Memory Bear, whose value depends not only on remembering more user history, but on using that history to provide more affect-sensitive, individualized, and contextually appropriate behavior over time. We hope this work serves as a useful step toward a richer evaluation ecosystem for affective memory in long-horizon interaction, including future extensions to broader scenarios, richer multimodal evidence, more realistic interaction settings, and stronger comparison across different memory architectures.

\newpage
\bibliography{MER_cite}
\bibliographystyle{plain}

\end{document}